\documentclass[runningheads]{llncs}
\usepackage[T1]{fontenc}
\usepackage{cite}
\usepackage{amssymb, amsmath, latexsym}
\usepackage{algorithm}
\usepackage{algpseudocode}
\usepackage{booktabs}
\usepackage{subcaption}
\usepackage{bm}
\usepackage{hyperref}
\usepackage{orcidlink}
\usepackage{color}

\makeatletter

\usepackage{wrapfig}
\usepackage{array}

\usepackage{graphicx}

\usepackage{url}  

\begin{document}
\title{Bootstrap The Original Latent: Learning a Private Model from a Black-box Model}
%


\titlerunning{Bootstrap The Original Latent}
%
\author{{Shuai Wang$^{*}$} \inst{1} \and
{Daoan Zhang$^{*}$}  \inst{2}\and
Jianguo Zhang\inst{2} \and
Weiwei Zhang\inst{3} \and
Rui Li\inst{1}
}

%
\authorrunning{}

%
\institute{Tsinghua University \and Southern University of Science and Technology \and Sinosmart Inc}

\footnotetext[1]{These authors contributed equally to this work.}
\maketitle              

\begin{abstract}
In this paper, considering the balance of data/model privacy of model owners and user needs, we propose a new setting called \textit{Back-Propagated Black-Box Adaptation (BPBA)} for users to better train their private models via the guidance of the back-propagated results of a Black-box foundation/source model. Our setting can ease the usage of the foundation/source models as well as prevent the leakage and misuse of foundation/source models. These phenomenons are more severe and common in medical image analysis. To better deal with the problems, we propose a new paradigm called \textit{Bootstrap The Original Latent (BTOL)} to fully utilize the foundation/source models. Our strategy consists of a trainable adapter and a freeze-and-thaw strategy. We apply our \textit{BTOL} under {BPBA} and Black-box UDA on three different medical image segmentation datasets. Experiments show that our paradigm is efficient and robust under various settings.
\end{abstract}

\section{Introduction}

Foundation Models such as ChatGPT \cite{ouyang2022training} and BEIT-3 \cite{wang2022image} are robust and efficient. They have an increasing power to generalize different types of data. It is a trend that people can adapt their private data by generating a tiny task-specific model under the instruction of the foundation models \cite{wang2022image}\cite{ouyang2022training}\cite{radford2021learning}. However, making such a foundation model available to everyone is a luxury. On the one hand, training such a big model is expensive even for big-name companies. On the other hand, the owners should take into account commercial considerations and data misuse. Therefore, foundation models such as GPT3\href{https://openai.com/product}{$^{1}$}, ERNIE 3.0 \cite{sun2021ernie} are only provided as a service when only black-box API can be accessed. 

Under the so-called Model-as-a-Service (MaaS) \cite{sun2022black} scenario, users can adapt their data via zero-shot or black-box source-free domain adaptation. These techniques work well on language tasks but are inefficient in visual tasks because the distribution gap in vision tasks are large \cite{zhang2023aggregation}. Even foundation vision models cannot generalize on such varieties of data. Thus, foundation models can help improve the efficiency of their task-specific models for the users\cite{wang2023ftso, wang2023Dionysus}, especially for those who only have little or unannotated data. However, it still requires a tuning process for users to adapt their data\cite{zeng2022simple,zeng2023substructure}.

The phenomenon and need in the medical image analysis scenarios are more severe and common \cite{mahmood2018unsupervised, zhang2020collaborative, guan2021domain}. The performance of the source model can be substantially degraded when transferred to a different distribution. Also, the need for data/model privacy protection in the hospital is more strict than in other scenarios\cite{wang2022mvsnet,wang2023flora}.

Thus, a common question in both MaaS and medical image analysis scenarios has emerged: \textbf{Can we balance the privacy of the foundation/source models and the practical needs for users to train their models?} A recently proposed strategy called black-box domain adaptation \cite{liang2022dine} provides a solution for model owners by only providing the forward propagated logits to users. In this paper, we extend this strategy to provide both the forward and backward propagated results. This operation can ease the users to adapt their data simultaneously without affecting the privacy of source data and models. This proposed paradigm is called \textit{Back-Propagated Black-Box Adaptation (BPBA)}. 

For a better illustration, as shown in Table. \ref{tab:paradigms}, conventional UDA \cite{ganin2015unsupervised} (Unsupervised Domain Adaptation) allows users to access both source data and model. White-box UDA\cite{liang2020we} bans the user's access to source data to preserve data privacy compared to conventional UDA. Black-box UDA \cite{zhang2021unsupervised} only allows the users to achieve the forward-propagation results (logits, etc.) to preserve the model privacy. While our BPBA additionally allows users to utilize the back-propagated information (usually gradients) compared to Black-box UDA. This relieves the burden for users to establish a tiny model on the private data as well as preserves the model privacy for source model owners.

\begin{table}[!t]
	\centering
	
	\caption{\textbf{Comparison of Different Existing Paradigms.} UDA indicates unsupervised domain adaptation; Param. Avail. indicates the availability or updating of the parameters of source model; For-propa. indicates forward-propagation; Back-propa. indicates back-propagation outputs.}
	\begin{center}
	   \begin{tabular}{|c|c|c|c|c|}
            \hline
                             & Source Data & Param. Avail. & For-propa. & Back-propa. \\
            \hline              
            Conventional UDA \cite{ganin2015unsupervised} & $\checkmark$ & $\checkmark$ & $\checkmark$ & $\checkmark$ \\
            White-box UDA \cite{liang2020we} & $\times$ & $\checkmark$ & $\checkmark$ & $\checkmark$ \\
            Black-box UDA \cite{zhang2021unsupervised} & $\times$ & $\times$ & $\checkmark$ & $\times$ \\
            BPBA (Ours) & $\times$ & $\times$ & $\checkmark$ & $\checkmark$ \\
            \hline
            \end{tabular}
    
	\end{center}
	\label{tab:paradigms}
    \vspace{-0.9cm}
\end{table}

\subsection{BPBA and Self-supervised learning}
To better understand the BPBA, we give an analysis and comparison of BPBA and self-supervised learning \cite{he2020momentum, chen2021exploring, grill2020bootstrap,zhang2022rethinking}. 
Since no source data and no ground truth are provided, we consider the adaption problem as a \textbf{conditional} self-supervised \textbf{alignment} problem. Both settings aim to learn good representations from the unlabeled data for the downstream tasks. Nevertheless, in BPBA, we have a conditional well-pre-trained model that contains lots of implicit information which can be inherited to optimize the target model. This is the crucial difference compared to the conventional self-supervised problem. How to leverage the hidden information in the source model shall be considered in the model design.

Moreover, take a brief look at traditional self-supervised learning; the critical properties of the success of self-supervised learning strategies is to maintain the alignment and uniformity of representations \cite{wang2020understanding}. The alignment indicates that the representations of the positive pair are close in the latent space. The uniformity indicates that the representations should be distributed as evenly as possible on the unit sphere. If we fail to align the positive pair, the model will learn the sub-optimal representations. If we lose uniformity, the model will suffer from mode collapse. In {BPBA}, we also need to consider both properties when adapting unlabeled data. However, the pre-trained model provides a strict mapping for unlabeled data, thus holding a stable and uniform distribution in the latent space. Therefore, other than unleashing the potential of the source model, another critical point to optimize the model is to build a better alignment strategy to align the unlabeled and invisible source data. 
\vspace{-0.2cm}

\subsection{Bootstrap The Original Latent}
To better utilize the user's data and guarantee the alignment, we propose a \textit{Freeze-and-thaw Adapter} strategy called \textit{BTOL} as shown in Fig. \ref{main}. Considering the potential domain gap between the unseen source data/model and target data, we design a novel cross-supervised paradigm to align the distribution via an adapter implicitly. To fully exploit the source model and unlabeled data, a freeze-and-thaw strategy is developed for the adapter and target model to teach each other. Notice that our \textit{BTOL} can also be utilized in the Black-box UDA task via a simulator as proposed in the \textit{lower} part of Fig. \ref{main}. Under both Black-box UDA and BPBA settings, our \textit{BTOL} can make full use of the source model and target data.

The contributions of this paper can be summarized into three folds:

(1) We find a better solution to balance the model privacy and user needs in both MaaS and medical image analysis by allowing users access to back-propagated information. We extend the Black-box UDA to Back-Propagated Black-Box Adaptation (BPBA) which can benefit both model owners and users.

(2) We propose a novel training strategy called \textit{BTOL} under the proposed setting. Our solution consists of an adapter and a freeze-and-thaw strategy. We then broaden our strategy to the Black-box UDA task, and our model also achieves state-of-the-art performance.

(3) Our paradigm outperforms all the Black-box UDA and White-box UDA methods on different settings and datasets.


\section{Methodology}
In this section, we introduce the novel approach \textit{BTOL} to better uncover the potential of the source model without any manual regularizations. For simplicity, in our method, adapters are used to narrow the distribution gap between the unseen source data and target data, freeze-and-thaw strategy can provide a constraint to avoid the representation collapse. To better exploit the potential of our method, we utilize our \textit{BTOL} in both the BPBA and black-box UDA settings. \textbf{Without} any manual augmentations, our method can easily outperform any existing black-box UDA methods and even some white-box UDA methods. Detailed methods and analysis are presented below.

\subsection{{BTOL} in the BPBA}
\begin{figure*}[!t]
	\centering
	\includegraphics[width=1.0\textwidth]{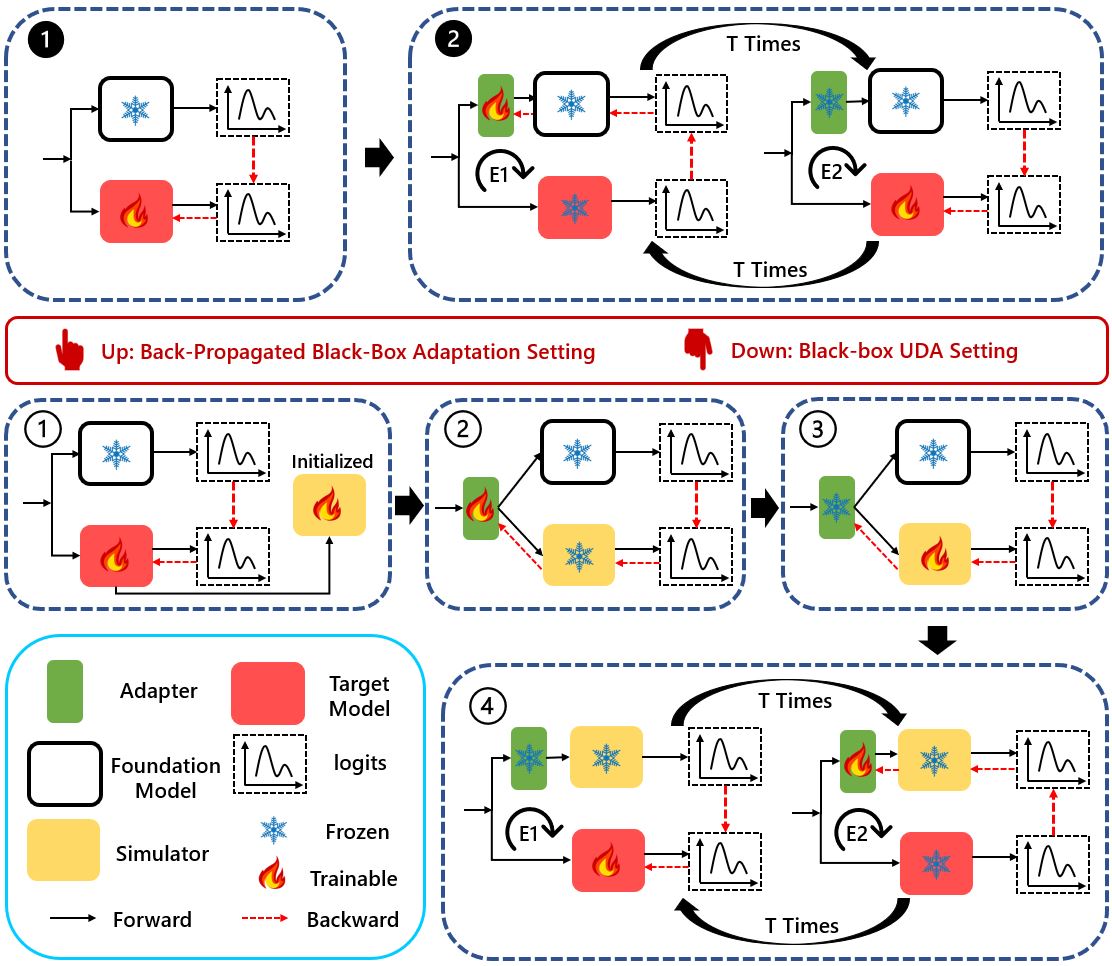}
	\caption{\textbf{Overview of our \textit{BTOL} under different settings.} \textbf{Upper: BPBA setting} ; \textbf{lower: Black-box UDA setting}}
    \label{main}
    \vspace{-0.4cm}
	
\end{figure*}

We first build the \textit{BTOL} strategy for BPBA, which is shown in the \textit{upper} part of Fig. \ref{main}. In BPBA, given a foundation/source model in the cloud server, which only allows users to achieve the output logits and back-propagated gradients. Users can utilize the two feed-backs to train their models. 

As the source model is efficient only on the source domain data, we aim to transfer the target domain to source-domain-like distribution to satisfy the source model. Thus the source model can produce a more accurate result. We set an adapter to deal with the domain transfer and align the distribution between the unseen source and target domains. 

We then introduce a dual EM\cite{moon1996expectation} approach called the \textit{freeze-and-thaw} strategy to deal with the alignment. In the conventional EM algorithm, the expectation step and maximization step are interleaved by recalculating the parameters and estimating the distribution. In our method, we make an extension and adaptation for the BPBA problem. The updating of the adapter and target model are interleaved by freezing one module and training another one. The pseudo-code is presented in Algorithm \ref{alg1}. 

The whole pipeline of our strategy is straightforward and easy to understand. We first trained the target model using the pseudo labels generated from source models. Then we train the adapter and target model via the freeze-and-thaw strategy. All the results will be reported in the experiments.

\subsection{{BTOL} in the Black-box UDA}

In the Black-box UDA setting, The back-propagated result is unavailable. Existing methods can only utilize manual augmentations, label-denoising techniques, and regularizations. They all fail to bootstrap the inner latent information in the original source model. Under this tougher condition, we introduce a simulator to simulate the source data distribution; then, we can apply our freeze-and-thaw under this setting.

The simulation step is presented in the second and third steps of the \textit{lower} part in Fig. \ref{main} after training the target model via the pseudo labels. We copy the weights of the target model to initialize the simulator. In the second step, we train the adapter to satisfy both the source and target models. This step helps the adapter output a mixture of source and target domains. In the third step, we freeze the adapter and train the simulator. This allows the simulator to get familiar with the output of the trained adapter, which is the mixed distribution of source and target domains. In the simulation procedure, the ideal distribution the simulator can imitate is the average mixture of the source and target distributions. After the simulation, we can use the trained simulator to replace the invisible source model to execute the freeze-and-thaw strategy.



\begin{algorithm}[t]

\caption{Training of \textit{BTOL} under BPBA.}\label{alg:cap}
\begin{algorithmic}[1]
\Require Random initialize adapter $A$ and target model $T$; Get access to source model $S$; The freeze-and-thaw loss is cross-entropy loss $L_{ce}$; $T$ is number of the training epoch. $E_1$ and $E_2$ are the numbers of tiny epoch in the training loop.

\State \textbf{Target Model Initialization:} Trained the target model $T$ with the pseudo labels generated by the source model $S$ until converged.

\State \textbf{Freeze-and-Thaw:}
\For{$t=0,1,\ldots,T-1$}
        
        \State \underline{target model $T$}:Freeze the parameters.
        \For{$e=0,1,\ldots,E_1-1$} 
        \State \underline{adapter $A$}: Learn parameters via the gradient downloaded from source model $S$ under the supervision from the pseudo output logits from target model $T$. 
        \EndFor

        \State \underline{adapter $A$}:Freeze the parameters.
         \For{$e=0,1,\ldots,E_2-1$} 
        \State \underline{target model $T$}: Learn parameters under the supervision from the pseudo output logits from the series connection of adapter $A$ and source model $S$.
        \EndFor
    
\EndFor
\end{algorithmic}
\label{alg1}

\end{algorithm}
\vspace{-0.4cm}




        

    

\section{Experiments}
\subsection{Dataset and Evaluation Metric.}
We evaluate our method on three tasks: fundus segmentation, cardiac structure segmentation, and prostate segmentation. For the fundus segmentation task, we use the training set of REFUGE challenge~\cite{orlando2020refuge} as the source domain and use the RIM-ONE-r3~\cite{fumero2011rim} as the target domain. Following~\cite{chen2021source}, we split the target domain into 99/60 images for training and testing, respectively. We resize each image to $512\times 512$ to feed the network. For the cardiac structure segmentation task, we choose the public ACDC dataset~\cite{BernardLZCYHCLC18} contains 200 volumes as the source domain. For the target domain, we use the LGE dataset from Multi-sequence Cardiac MR segmentation Challenge (MSCMR2019)~\cite{Zhuang19}. We split it into 80\%/20\% for training and testing. We use 2d slices for training, and all images are resized to $192\times 192$ as the network input. As for prostate segmentation, we use the MSD05~\cite{Antonelli2022} as the source domain and Promise12~\cite{litjens2014evaluation} as the target domain. We also split it into 80\%/20\% for training and testing. For evaluation, we use two commonly-used metrics in the medical segmentation field: Dice Score (DSC) and Average Surface Distance (ASD). DSC measures the overlap between prediction and ground truth, and ASD represents the performance at the object boundary. Higher DSC and lower ASD mean better performance.

\begin{table}[!h]
	\centering
 \vspace{-0.8cm}
	
	\caption{Experiments of fundus segmentation. The source model is trained on REFUGE challenge and the target data is RIM-ONE-r3. The baseline model is the target model which is trained on the pseudo label. Type "W" indicates the method is a white-box UDA method; the Type "B" indicates BTOL in the black-box UDA; Type "P" indicates BTOL in the BPBA.}
	\begin{center}
	   \begin{tabular}{|c|c|c|c|c|c|c|c|}
            \hline
            Methods & Type & \multicolumn{3}{|c|}{Dice $\uparrow$} & \multicolumn{3}{|c|}{ASD$\downarrow$} \\
            \hline
                           &  & Disc & Cup & Avg. & Disc & Cup & Avg. \\
            \hline              
            Source Model &- & $88.34 ^{\pm 4.48}$ & $71.35 ^{\pm 22.75}$ & $79.85 ^{\pm 12.59} $ & $10.65 ^{\pm 4.27}$  & $10.75 ^{\pm 5.34}$ & $10.7 ^{\pm 4.18}$ \\ 
            Baseline Model  & -& $91.34 ^{\pm 4.49}$ & $71.85 ^{\pm 17.00}$ & $81.59 ^{\pm 9.83}$ & $7.62 ^{\pm 3.65}$ & $10.64 ^{\pm 6.71}$ & $9.00 ^{\pm 3.87}$ \\
            SRDA \cite{bateson2020source} & W & $89.37 ^{\pm 2.70}$ & $77.61 ^{\pm 13.58}$ & $83.49 ^{\pm 8.14}$ & $9.91 ^{\pm 2.45}$ & $10.15 ^{\pm 5.75}$ & $10.03 ^{\pm 4.10}$ \\
            AdvEnt \cite{vu2019advent} & W & $89.73 ^{\pm 3.66}$ & $77.99 ^{\pm 21.08}$ & $83.86 ^{\pm 12.37}$ & $9.84 ^{\pm 3.86}$ & $7.57 ^{\pm 4.24}$ & $8.71^{\pm 8.10}$ \\
            DAE \cite{karani2021test} & W & $89.08 ^{\pm 3.32}$ & $79.01 ^{\pm 12.82}$ & $84.41 ^{\pm 8.07}$ & $11.63 ^{\pm 6.84}$ & $10.31 ^{\pm 8.45}$ & $10.97^{\pm 7.65}$ \\
            DPL \cite{chen2021source} & W & $90.13 ^{\pm 3.06}$ & $79.78 ^{\pm 11.05}$ & $84.95 ^{\pm 7.06}$ & $9.43 ^{\pm 3.46}$ & $9.01 ^{\pm 5.59}$ & $9.22^{\pm 4.53}$ \\
            \hline
            EMD \cite{liu2022unsupervised} & B & $90.50 ^{\pm 3.78}$ & $73.50 ^{\pm 11.56}$ & $82.00 ^{\pm 8.76}$ & $10.52 ^{\pm 4.18}$ & $7.12 ^{\pm 4.15}$ & $8.82^{\pm 2.59}$ \\
            \textit{BTOL}(Ours) & B & $92.54 ^{\pm 4.08}$ & $72.86 ^{\pm 13.46}$ & $82.70 ^{\pm 8.24}$ & $8.45 ^{\pm 3.32}$ & $9.14 ^{\pm 3.64}$ & $8.80^{\pm 3.02}$ \\
            \textit{BTOL}(Ours) & P & $91.52 ^{\pm 4.19}$ & $78.73 ^{\pm 11.87}$ & $\mathbf{85.13 ^{\pm 7.14}}$ & $7.53 ^{\pm 3.36}$ & $8.76 ^{\pm 3.66}$ & $\mathbf{8.14^{\pm 2.74}}$ \\
            \hline
            \end{tabular}
    
	\end{center}
	\label{tab:fundus}
\end{table}

\begin{table}[!h]
	\centering

	\caption{Experiments of cardiac structure segmentation and prostate segmentation. Due to the page limitation, we choose the previous sota method DPL for a fair comparison.}
	\begin{center}
	   \begin{tabular}{|c|c|c|c|c|c|}
            \hline
             \multicolumn{6}{|c|}{\textbf{Cardiac Structure Segmentation}}  \\
             \hline
             Methods & Type & \multicolumn{4}{|c|}{Dice $\uparrow$}  \\
            \hline
                           &  & RV & Myo & LV & Avg.\\
            \hline              
            Source Model &- & $40.28 ^{\pm 26.73}$ & $48.83 ^{\pm 10.84}$ & $76.45 ^{\pm 10.21} $ & $55.19 ^{\pm 14.19}$   \\ 
            Baseline Model  & -& $46.67 ^{\pm 30.49}$ & $53.95 ^{\pm 9.70}$ & $76.01 ^{\pm 9.11}$ & $58.87 ^{\pm 13.37}$  \\
            DPL \cite{chen2021source} & W & $48.14^{\pm 29.45}$	& $53.76^{\pm 9.11}$ & $ 78.42^{\pm 8.64}$  & 	$ 60.11^{\pm 15.73}$\\
            \hline
            EMD \cite{liu2022unsupervised} & B & $47.59 ^{\pm 28.46}$ & $53.67 ^{\pm 9.79}$ & $75.48 ^{\pm 9.58}$ & $58.91 ^{\pm 13.48}$  \\
            
            \textit{BTOL}(Ours) & B & $47.12 ^{\pm 29.45}$ & $53.85 ^{\pm 10.15}$ & $78.45 ^{\pm 9.66}$ & $59.81 ^{\pm 12.86}$ \\
            \textit{BTOL}(Ours) & P & $49.78^{\pm 25.45}$ & $54.12 ^{\pm 9.46}$ & $76.52 ^{\pm 10.40}$ & $\mathbf{60.14 ^{\pm 13.04}}$ \\
            \hline
            \hline
            \multicolumn{6}{|c|}{\textbf{Cardiac Structure Segmentation}}  \\
             
            \hline
            Methods & Type & \multicolumn{4}{|c|}{ASD $\downarrow$}  \\
            \hline
                           &  & RV & Myo & LV & Avg.\\
            \hline              
            Source Model &- & $4.50 ^{\pm 3.42}$ & $4.60 ^{\pm 2.51}$ & $5.78 ^{\pm 2.02} $ & $4.96 ^{\pm 1.74}$   \\ 
            Baseline Model  & -& $2.03 ^{\pm 1.61}$ & $4.14 ^{\pm 1.87}$ & $5.52 ^{\pm 2.36}$ & $3.90 ^{\pm 1.28}$  \\
            DPL \cite{chen2021source} & W & $1.55 ^{\pm 1.24}$ & $4.75 ^{\pm 2.04}$ & $4.95 ^{\pm 2.23}$ & $3.75 ^{\pm 1.20}$  \\
            \hline
            EMD \cite{liu2022unsupervised} & B & $2.12 ^{\pm 1.47}$ & $4.25 ^{\pm 1.95}$ & $5.13 ^{\pm 2.78}$ & $3.83 ^{\pm 1.41}$  \\
            \textit{BTOL}(Ours) & B & $1.76 ^{\pm 1.45}$ & $4.55 ^{\pm 1.34}$ & $5.32 ^{\pm 2.78}$ & $3.88 ^{\pm 1.24}$ \\
            \textit{BTOL}(Ours) & P & $1.32^{\pm 1.11}$ & $4.24 ^{\pm 1.92}$ & $5.45 ^{\pm 2.18}$ & $\mathbf{3.67 ^{\pm 1.14}}$ \\
            \hline
            \end{tabular}
    
            \begin{tabular}{|c|c|c|c|}
            \hline
            \multicolumn{4}{|c|}{\textbf{Prostate Segmentation}}  \\
          
            \hline
            Methods & Type & Dice $\uparrow$ & ASD $\downarrow$ \\
            \hline              
            Source Model &- & $47.50 ^{\pm 26.21}$ & $9.80 ^{\pm 8.84}$    \\ 
            Baseline Model  & -& $51.68 ^{\pm 24.56}$ & $8.99 ^{\pm 8.25}$  \\
            DPL \cite{chen2021source} & W & $52.95^{\pm 23.14}$ & $7.98^{\pm7.26}$ \\
            \hline
            EMD \cite{liu2022unsupervised} & B & $52.47 ^{\pm 23.18}$ & $8.11 ^{\pm 7.85}$ \\
            
            \textit{BTOL}(Ours) & B & $52.40 ^{\pm 23.18}$ & $7.88 ^{\pm 7.02}$ \\
            \textit{BTOL}(Ours) & P & $\mathbf{54.16^{\pm 22.75}}$ & $\mathbf{6.58 ^{\pm 6.98}}$  \\
            \hline
            \end{tabular}
    
	\end{center}
	\label{tab:cardiac}
  \vspace{-0.7cm}
\end{table}

\begin{table}[!h]
	\centering
	
	\caption{Experiments of fundus segmentation of different backbones. The source model is DeepLabV3+ and target model is UNet\cite{ronneberger2015u}. The source model is trained on REFUGE challenge and the target data is RIM-ONE-r3. Experiments show that our method can work on different target model backbones.}
	\begin{center}
	   \begin{tabular}{|c|c|c|c|c|c|c|c|}
            \hline
            Methods & Type & \multicolumn{3}{|c|}{Dice $\uparrow$} & \multicolumn{3}{|c|}{ASD$\downarrow$} \\
            \hline
                           &  & Disc & Cup & Avg. & Disc & Cup & Avg. \\
            \hline              
        
            Baseline Model  & -& $46.39 ^{\pm 34.86}$ & $75.06 ^{\pm 27.98}$ & $60.73 ^{\pm 29.42}$ & $14.90 ^{\pm 11.52}$ & $15.61 ^{\pm 14.45}$ & $15.26 ^{\pm 8.95}$ \\
            \hline
            \textit{BTOL}(Ours) & B & $47.40 ^{\pm 31.20}$ & $76.48 ^{\pm 26.55}$ & $61.94 ^{\pm 28.47}$ & $14.56 ^{\pm 10.76}$ & $13.00 ^{\pm 12.45}$ & $13.78^{\pm 8.62}$ \\
            \textit{BTOL}(Ours) & P & $49.70 ^{\pm 32.68}$ & $78.15 ^{\pm 26.80}$ & $\mathbf{63.93 ^{\pm 27.67}}$ & $13.56 ^{\pm 9.80}$ & $12.17 ^{\pm 10.64}$ & $\mathbf{12.87^{\pm 9.15}}$ \\
            \hline
            \end{tabular}
    
	\end{center}
	\label{tab:abla}
 \vspace{-0.7cm}
\end{table}

\subsection{Implementation Details}
For all experiments, we use DeepLabV3+~\cite{deeplabv3+} with MobileNetV2~\cite{mobilenetv2} backbone as the segmentation model (e.g., Target model or Simulator in Figure~\ref{main}). For the adapter, we use a neural network consisting of three blocks, and each block includes a 2D convolution layer and Relu activation function. We use Adam optimizer with learning rate as $1e^{-4}$ and set batch size as 8 without specific choice. We train the target model from scratch 100 epochs in the target model initialization stage. After that, we set $T=4$, $E_1=10$ and $E_2=30$ in Figure \ref{main}.

\subsection{Experimental Results}
We test our method mainly on medical image datasets from different body parts to prove efficacy and robustness, including fundus segmentation, cardiac structure segmentation, and prostate segmentation.
As presented in Table. \ref{tab:cardiac}, \ref{tab:fundus}, When applying \textit{BTOL} to black-box UDA tasks, our model can easily outperform previous methods \textbf{without} any augmentations, which means our method can fully exploit the potential information from the source model. When applying our method to the BPBA setting without getting access to or updating the source model parameters, our model outperforms all white-box UDA methods under all the settings. This proves that our model can make more efficient use of the source model, even with fewer operations. We believe that if adding more augmentations and regularizations, our method can achieve more impressive results.

We also test out strategy by changing the backbone of the target model as shown in \ref{tab:abla}. We changed the target model backbone from DeepLabV3+ to UNet\cite{ronneberger2015u}. Our strategy can still stably outperform the baseline model, which shows our strategy is robust and easy to use in various types of backbone.

\vspace{-0.15cm}
\section{Discussion}

From the view of self-supervised learning, our \textit{BTOL} can be thought of as a conditional dual SimSiam \cite{chen2021exploring}. SimSiam attributes the collapsing solutions to the stop-gradient operation. In our method, since there is a well-trained source model, we can thus broaden the unilateral stop-gradient operation into the dual freeze-and-thaw strategy without considering the model collapse. This EM-like approach can adequately utilize the source model. Moreover, our model does not need a pair of augmented inputs like SimSiam because the source model can provide a fixed distribution to maintain uniformity. If adding more augmentation procedures like SimSiam, we believe our model can achieve an improvement.  

\vspace{-0.15cm}
\section{Conclusion}
In this paper, we address a new setting, BPBA, to deal with the balance between model privacy and user needs when adapting the Black-box foundation/source models. Compared to Black-box UDA, we release the back-propagated information for users to train their models. We also propose a new strategy called \textit{BTOL} to fully utilize the implicit information from the source model to optimize the target model. \textit{BTOL} contains an adapter and a freeze-and-thaw strategy to cross-teach the adapter and target model. Our paradigm outperforms all the methods on different settings. Our setting and solution can benefit nearly all the fields which utilize the foundation/source models. We hope our work can provide a new perspective and solution in the new era of foundation models.
\bibliographystyle{splncs04.bst}
\bibliography{reference}

\begin{thebibliography}{10}
\providecommand{\url}[1]{\texttt{#1}}
\providecommand{\urlprefix}{URL }
\providecommand{\doi}[1]{https://doi.org/#1}

\bibitem{Antonelli2022}
Antonelli, M., Reinke, A., Bakas, S., Farahani, K., et~al.: The medical
  segmentation decathlon. Nature Communications  \textbf{13}(1) (Jul 2022)

\bibitem{bateson2020source}
Bateson, M., Kervadec, H., Dolz, J., Lombaert, H., Ayed, I.B.: Source-relaxed
  domain adaptation for image segmentation. In: International Conference on
  Medical Image Computing and Computer-Assisted Intervention. pp. 490--499.
  Springer (2020)

\bibitem{BernardLZCYHCLC18}
Bernard, O., Lalande, A., Zotti, C., et~al.: Deep learning techniques for
  automatic mri cardiac multi-structures segmentation and diagnosis: is the
  problem solved? IEEE Transactions on Medical Imaging  \textbf{37}(11),
  2514--2525 (2018)

\bibitem{chen2021source}
Chen, C., Liu, Q., Jin, Y., Dou, Q., Heng, P.A.: Source-free domain adaptive
  fundus image segmentation with denoised pseudo-labeling. In: MICCAI. pp.
  225--235. Springer (2021)

\bibitem{deeplabv3+}
Chen, L., Zhu, Y., Papandreou, G., Schroff, F., Adam, H.: Encoder-decoder with
  atrous separable convolution for semantic image segmentation. In: ECCV (2018)

\bibitem{chen2021exploring}
Chen, X., He, K.: Exploring simple siamese representation learning. In:
  Proceedings of the IEEE/CVF conference on computer vision and pattern
  recognition. pp. 15750--15758 (2021)

\bibitem{fumero2011rim}
Fumero, F., Alay{\'o}n, S., Sanchez, J.L., Sigut, J., Gonzalez-Hernandez, M.:
  Rim-one: An open retinal image database for optic nerve evaluation. In:
  international symposium on computer-based medical systems. pp.~1--6. IEEE
  (2011)

\bibitem{ganin2015unsupervised}
Ganin, Y., Lempitsky, V.: Unsupervised domain adaptation by backpropagation.
  In: International conference on machine learning. pp. 1180--1189. PMLR (2015)

\bibitem{grill2020bootstrap}
Grill, J.B., Strub, F., Altch{\'e}, F., Tallec, C., Richemond, P., Buchatskaya,
  E., Doersch, C., Avila~Pires, B., Guo, Z., Gheshlaghi~Azar, M., et~al.:
  Bootstrap your own latent-a new approach to self-supervised learning.
  Advances in neural information processing systems  \textbf{33},  21271--21284
  (2020)

\bibitem{guan2021domain}
Guan, H., Liu, M.: Domain adaptation for medical image analysis: a survey. IEEE
  Transactions on Biomedical Engineering  \textbf{69}(3),  1173--1185 (2021)

\bibitem{he2020momentum}
He, K., Fan, H., Wu, Y., Xie, S., Girshick, R.: Momentum contrast for
  unsupervised visual representation learning. In: Proceedings of the IEEE/CVF
  conference on computer vision and pattern recognition. pp. 9729--9738 (2020)

\bibitem{karani2021test}
Karani, N., Erdil, E., Chaitanya, K., Konukoglu, E.: Test-time adaptable neural
  networks for robust medical image segmentation. MIA  \textbf{68},  101907
  (2021)

\bibitem{liang2020we}
Liang, J., Hu, D., Feng, J.: Do we really need to access the source data?
  source hypothesis transfer for unsupervised domain adaptation. In:
  International Conference on Machine Learning. pp. 6028--6039. PMLR (2020)

\bibitem{liang2022dine}
Liang, J., Hu, D., Feng, J., He, R.: Dine: Domain adaptation from single and
  multiple black-box predictors. In: Proceedings of the IEEE/CVF Conference on
  Computer Vision and Pattern Recognition. pp. 8003--8013 (2022)

\bibitem{litjens2014evaluation}
Litjens, G., Toth, R., van~de Ven, W., Hoeks, C., Kerkstra, S., van Ginneken,
  B., Vincent, G., Guillard, G., Birbeck, N., Zhang, J., et~al.: Evaluation of
  prostate segmentation algorithms for mri: the promise12 challenge. Medical
  image analysis  \textbf{18}(2),  359--373 (2014)

\bibitem{liu2022unsupervised}
Liu, X., Yoo, C., Xing, F., Kuo, C.C.J., El~Fakhri, G., Kang, J.W., Woo, J.:
  Unsupervised black-box model domain adaptation for brain tumor segmentation.
  Frontiers in Neuroscience p.~341 (2022)

\bibitem{mahmood2018unsupervised}
Mahmood, F., Chen, R., Durr, N.J.: Unsupervised reverse domain adaptation for
  synthetic medical images via adversarial training. IEEE transactions on
  medical imaging  \textbf{37}(12),  2572--2581 (2018)

\bibitem{moon1996expectation}
Moon, T.K.: The expectation-maximization algorithm. IEEE Signal processing
  magazine  \textbf{13}(6),  47--60 (1996)

\bibitem{orlando2020refuge}
Orlando, J.I., Fu, H., Breda, J.B., van Keer, K., Bathula, D.R., et~al.: Refuge
  challenge: A unified framework for evaluating automated methods for glaucoma
  assessment from fundus photographs. Medical image analysis  \textbf{59},
  101570 (2020)

\bibitem{ouyang2022training}
Ouyang, L., Wu, J., Jiang, X., Almeida, D., Wainwright, C.L., Mishkin, P.,
  Zhang, C., Agarwal, S., Slama, K., Ray, A., et~al.: Training language models
  to follow instructions with human feedback. arXiv preprint arXiv:2203.02155
  (2022)

\bibitem{radford2021learning}
Radford, A., Kim, J.W., Hallacy, C., Ramesh, A., Goh, G., Agarwal, S., Sastry,
  G., Askell, A., Mishkin, P., Clark, J., et~al.: Learning transferable visual
  models from natural language supervision. In: International conference on
  machine learning. pp. 8748--8763. PMLR (2021)

\bibitem{ronneberger2015u}
Ronneberger, O., Fischer, P., Brox, T.: U-net: Convolutional networks for
  biomedical image segmentation. In: International Conference on Medical image
  computing and computer-assisted intervention. pp. 234--241. Springer (2015)

\bibitem{mobilenetv2}
Sandler, M., Howard, A.G., Zhu, M., Zhmoginov, A., Chen, L.: Mobilenetv2:
  Inverted residuals and linear bottlenecks. In: CVPR (2018)

\bibitem{sun2022black}
Sun, T., Shao, Y., Qian, H., Huang, X., Qiu, X.: Black-box tuning for
  language-model-as-a-service. In: International Conference on Machine
  Learning. pp. 20841--20855. PMLR (2022)

\bibitem{sun2021ernie}
Sun, Y., Wang, S., Feng, S., Ding, S., Pang, C., Shang, J., Liu, J., Chen, X.,
  Zhao, Y., Lu, Y., et~al.: Ernie 3.0: Large-scale knowledge enhanced
  pre-training for language understanding and generation. arXiv preprint
  arXiv:2107.02137  (2021)

\bibitem{vu2019advent}
Vu, T.H., Jain, H., Bucher, M., et~al.: Advent: Adversarial entropy
  minimization for domain adaptation in semantic segmentation. In: CVPR. pp.
  2517--2526 (2019)

\bibitem{wang2023Dionysus}
Wang, L.: Dionysus: Recovering scene structures by dividing into semantic
  pieces

\bibitem{wang2023ftso}
Wang, L., Chen, L.: Ftso: Effective nas via first topology second operator.
  arXiv preprint arXiv:2303.12948  (2023)

\bibitem{wang2022mvsnet}
Wang, L., Gong, Y., Ma, X., Wang, Q., Zhou, K., Chen, L.: Is-mvsnet: Importance
  sampling-based mvsnet. In: Computer Vision--ECCV 2022: 17th European
  Conference, Tel Aviv, Israel, October 23--27, 2022, Proceedings, Part XXXII.
  pp. 668--683. Springer (2022)

\bibitem{wang2023flora}
Wang, L., Gong, Y., Wang, Q., Zhou, K., Chen, L.: Flora: dual-frequency
  loss-compensated real-time monocular 3d video reconstruction. In: Proceedings
  of the AAAI Conference on Artificial Intelligence. vol.~1 (2023)

\bibitem{wang2020understanding}
Wang, T., Isola, P.: Understanding contrastive representation learning through
  alignment and uniformity on the hypersphere. In: International Conference on
  Machine Learning. pp. 9929--9939. PMLR (2020)

\bibitem{wang2022image}
Wang, W., Bao, H., Dong, L., Bjorck, J., Peng, Z., Liu, Q., Aggarwal, K.,
  Mohammed, O.K., Singhal, S., Som, S., et~al.: Image as a foreign language:
  Beit pretraining for all vision and vision-language tasks. arXiv preprint
  arXiv:2208.10442  (2022)

\bibitem{zeng2023substructure}
Zeng, D., Liu, W., Chen, W., Zhou, L., Zhang, M., Qu, H.: Substructure aware
  graph neural networks. In: Proc. of AAAI (2023)

\bibitem{zeng2022simple}
Zeng, D., Zhou, L., Liu, W., Qu, H., Chen, W.: A simple graph neural network
  via layer sniffer. In: ICASSP 2022-2022 IEEE International Conference on
  Acoustics, Speech and Signal Processing (ICASSP). pp. 5687--5691. IEEE (2022)

\bibitem{zhang2023aggregation}
Zhang, D., Chen, M., Li, C., Huang, L., Zhang, J.: Aggregation of
  disentanglement: Reconsidering domain variations in domain generalization.
  arXiv preprint arXiv:2302.02350  (2023)

\bibitem{zhang2022rethinking}
Zhang, D., Li, C., Li, H., Huang, W., Huang, L., Zhang, J.: Rethinking
  alignment and uniformity in unsupervised image semantic segmentation. arXiv
  preprint arXiv:2211.14513  (2022)

\bibitem{zhang2021unsupervised}
Zhang, H., Zhang, Y., Jia, K., Zhang, L.: Unsupervised domain adaptation of
  black-box source models. arXiv preprint arXiv:2101.02839  (2021)

\bibitem{zhang2020collaborative}
Zhang, Y., Wei, Y., Wu, Q., Zhao, P., Niu, S., Huang, J., Tan, M.:
  Collaborative unsupervised domain adaptation for medical image diagnosis.
  IEEE Transactions on Image Processing  \textbf{29},  7834--7844 (2020)

\bibitem{Zhuang19}
Zhuang, X.: Multivariate mixture model for myocardial segmentation combining
  multi-source images. IEEE Transactions on Pattern Analysis and Machine
  Intelligence  \textbf{41}(12),  2933--2946 (2018)

\end{thebibliography}
\clearpage
\begin{figure}[!h]
    \centering
    \includegraphics[width=0.8\textwidth]{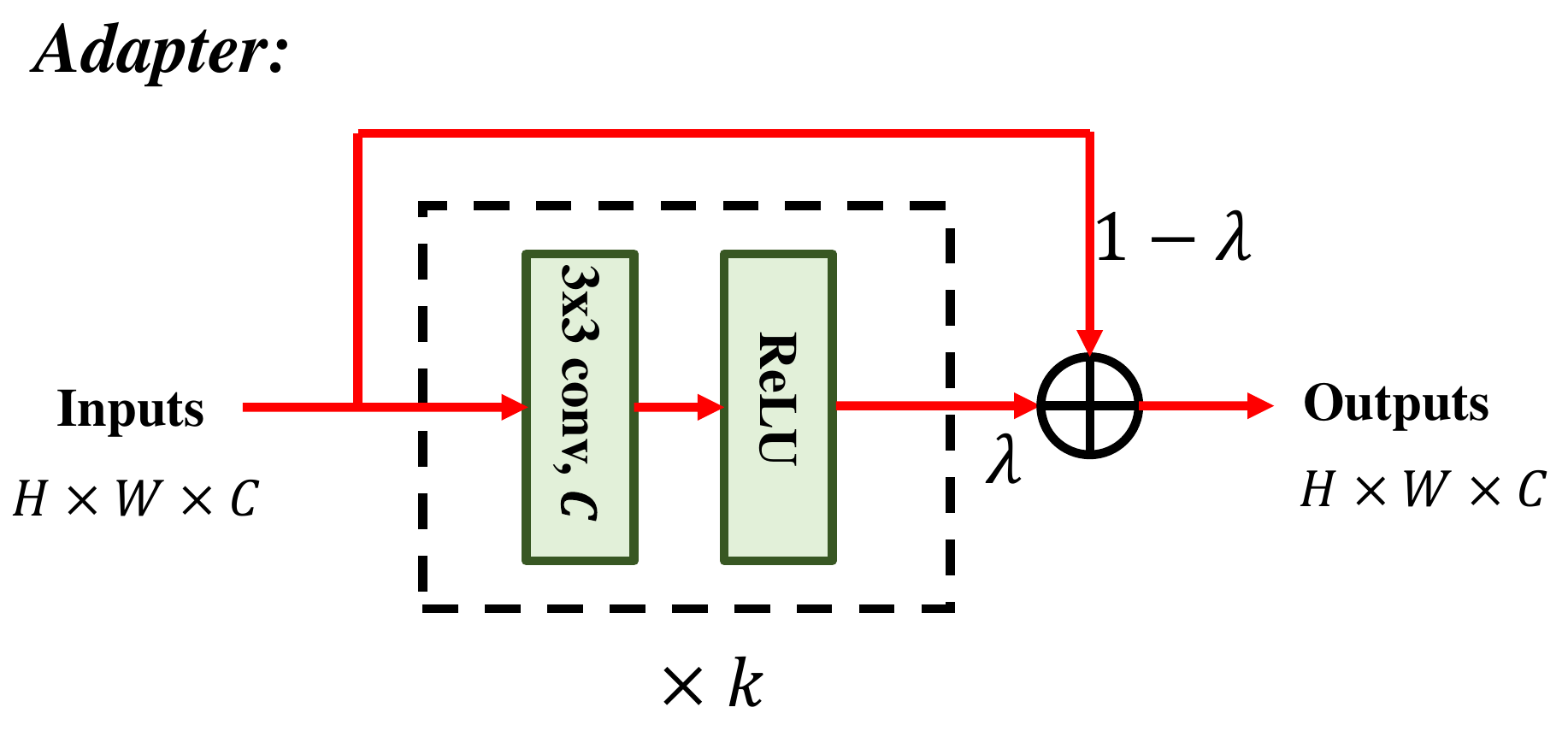}
    \caption{The structure of Adapter in Figure \ref{main}. The Adapter consists of $k$ blocks and a residual skip connection. We set $k=3$ in all experiments. $H$,$W$,$C$ denote the height, width and the number of channels of inputs, respectively.}
    \label{fig:adapter}
\end{figure}
\begin{table}[!h]
    \centering
    \setlength{\tabcolsep}{1.5mm}
    \caption{Results on different $k$ (the number of blocks in Adapter) on the RIM-ONE-r3 dataset.}
    \begin{tabular}{cccc}
    \toprule
        $k$ & 1 & 3 & 5  \\
        \hline
         Dice (\%) & 83.19$\pm$8.56 & \textbf{85.13}$\pm$7.14 & 84.92$\pm$7.56 \\
    \bottomrule
    \end{tabular}   
    \label{tab:ablation_k}
\end{table}

\end{document}


\title{Bootstrap The Original Latent: Learning a Private Model from a Black-box Model}
%


\titlerunning{Bootstrap The Original Latent}
%
\author{}

%
\authorrunning{}

%
\institute{}

\renewcommand{\thefootnote}{\fnsymbol{footnote}}
%
\maketitle              

\begin{figure}[!h]
    \centering
    \includegraphics[width=0.8\textwidth]{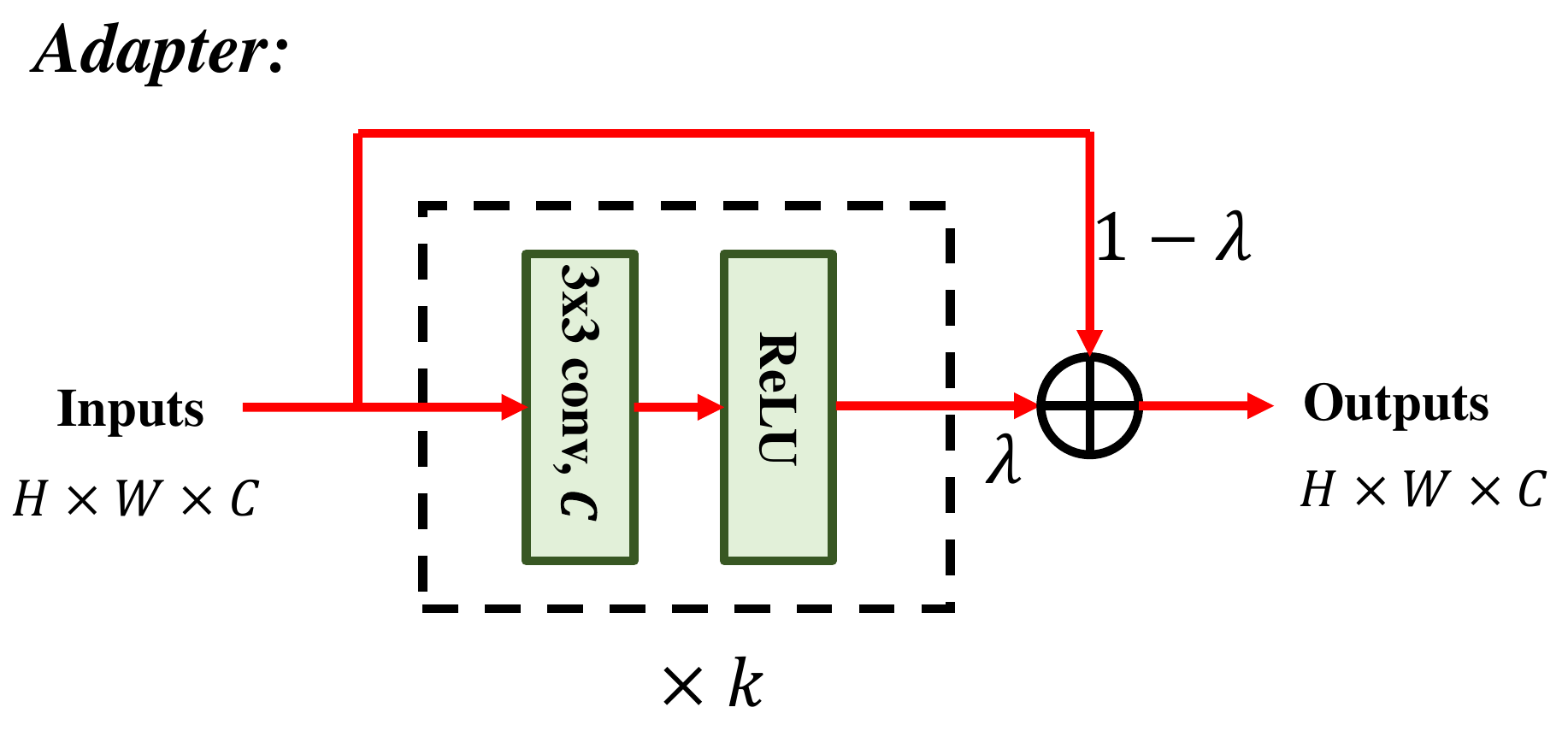}
    \caption{The structure of the adapter. The adapter consists of $k$ blocks and a residual skip connection. We set $k=3$ in all the experiments. $H$, $W$, $C$ denote the height, width and the number of channels of the inputs, respectively.}
    \label{fig:adapter}
\end{figure}

We first present the detailed structure of adapter in \ref{fig:adapter}, which is consist of a conv block, a ReLU function and a skip connection. We then ablates the number of blocks in the adapter chosen from {1, 3, 5}. The results is presented in \ref{tab:ablation_k}. Results show that with $k=3$, the whole model can achieve a top performance on RIM-ONE-r3 dataset.

\begin{table}[!h]
    \centering
    \setlength{\tabcolsep}{1.5mm}
    \caption{Results on different $k$ (the number of blocks in the adapter) on the RIM-ONE-r3 dataset.}
    \begin{tabular}{cccc}
    \toprule
        $k$ & 1 & 3 & 5  \\
        \hline
         Dice (\%) & 83.19$\pm$8.56 & \textbf{85.13}$\pm$7.14 & 84.92$\pm$7.56 \\
    \bottomrule
    \end{tabular}   
    \label{tab:ablation_k}
\end{table}

\bibliographystyle{splncs04.bst}